# Combination of Domain Knowledge and Deep Learning for Sentiment Analysis

Khuong Vo[1], Dang Pham[1], Mao Nguyen[1], Trung Mai[2]
and Tho Quan[2]

[1] YouNet Corporation - 2nd floor, Lu Gia Plaza, 70 Lu Gia Street, District 11,
Ho Chi Minh City
{khuongva, dangpnh, maonx}@younetco.com
[2] Bach Khoa University - 268 Ly Thuong Kiet Street, District 10,
Ho Chi Minh City
{mdtrung, qttho}@hcmut.edu.vn

**Abstract.** The emerging technique of deep learning has been widely applied in many different areas. However, when adopted in a certain specific domain, this technique should be combined with domain knowledge to improve efficiency and accuracy. In particular, when analyzing the applications of deep learning in sentiment analysis, we found that the current approaches are suffering from the following drawbacks: (i) the existing works have not paid much attention to the importance of different types of sentiment terms, which is an important concept in this area; and (ii) the loss function currently employed does not well reflect the degree of error of sentiment misclassification. To overcome such problem, we propose to combine domain knowledge with deep learning. Our proposal includes using sentiment scores, learnt by quadratic programming, to augment training data; and introducing *penalty matrix* for enhancing the loss function of cross entropy. When experimented, we achieved a significant improvement in classification results.

## 1 Introduction

*Opinion* was defined by Oxford Dictionary as the feeling or the thought of someone about something and these thoughts are not necessarily the truth. Therefore, opinion is always an important reference for making decisions of individuals and organizations. Before the Internet, opinions were referenced via friends, family or consumer opinion polls of enterprises. The explosion of information and communication technologies (ICT) leads to a huge amount of information to be read. Some information is quite "big" but not containing much useful information. This causes difficulties for individuals and businesses in consulting, searching, synthesizing information as well as evaluating and tracking customer comments on the products and services of the business. Therefore, *opinion mining/sentiment analysis* has been born and is developed rapidly, strongly and attracting much attention in research communities. According to Liu [1], opinion has a significant role in daily activities of people due to the fact that important decision is proposed from the consultation of the others.

Research on this topic was conducted at different levels: *term level* [2], *phrase level* [3], *sentence level* [4] and *document level* [5,6]. In terms of methodologies, approaches related to this problem can be summarized as follows:

- *Lexicon approach*: *sentiments terms* are used a lot in sentiment analysis. There are *positive* terms and *negative* terms. Additionally, there are also opinion phrases or idioms, which can be grouped into *Opinion Lexicon* [7]. Dictionary-based method by *Minging* and *Kim* [8, 9] shows strategies using dictionary for identifying sentiment terms.
- *Corpus-based methods*: *This* method is based on syntax and pattern analysis to find sentiment words in a big dataset [10].

Recently, with the introduction of *TreeBank*, especially *Stanford Sentiment Treebank* [11], sentiment analysis using *deep learning* becomes an emerging trend in the field. Recursive Neural Tensor Network (RNTN) was applied to the treebank and produced high performance [11]. Formerly, the compositionality idea related to neural networks has been discussed by Hinton [12], and the idea of feeding a neural network with inputs through multiple-way interactions, parameterized by a tensor have been proposed for relation classification [13]. Along with the treebank, the famous *Stanford CoreNLP tool* [14] is used widely by the community for sentiment tasks. Besides, convolution-based method continues to be developed for sentiment analysis on sentences [15, 16]. To store occurrence order relationship between features, recurrent neural network systems such as Long Short Term Memory (LSTM) was used in combination with convolution to perform sentiment analysis for short text [17]. Most recently, a combined architecture using deep learning for sentiment analysis has been proposed in [18].

However, when deep learning is used with real datasets from different sentiment domains, we observe that there are some problems arising as follows:

- Each domain has a different set of sentiment terms. For example, for *Smartphone*, positive/negative terms can be *durable, expensive, well-designed, slim,* etc. Meanwhile, in *Airlines*, sentiment terms can be *delay, slow check-in, good service*, etc. Each term carries a different *sentiment score*. To date, these aspects seem not considered much in deep learning approaches.
- Like other neural networks, deep learning uses a *loss function* to evaluate the error of the learning process. Currently, for sentiment analysis approaches, the default loss function assigns the same error rate for different error cases. For example, if a training sample is expected as a *negative* case, the loss function will produce the same error value if this sample is wrongly predicted as positive or neutral cases. Intuitively, misclassification from *negative to positive* should be considered more serious than from *negative to neutral*. We believe that the loss function should assign different values for those cases.

To tackle these problems, we propose the following approaches:

- We use *quadratic programming* [24] to learn sentiment scores for sentiment terms. Then, we use these sentiment scores to perform *augmentation* of the dataset to train deep learning models.
- We improve the loss function of the deep learning model by applying a *penalty matrix* so that the system can learn more accurately from errors of different misclassification cases.

The rest of this paper is organized as follows. In Section 2, we recall some background on convolution neural networks (CNN) and sentiment analysis. A general architecture of using CNN for sentiment analysis is presented in Section 3. Section 4 shows the contribution of our study about learning sentiment scores using quadratic programming and how this score is used for data augmentation. Section 5 discusses the idea of using a *penalty matrix* to improve the loss function. Section 6 presents the results of our experiments. Finally, Section 7 concludes the paper.

## 2  Background

### 2.1  Convolution Neural Networks (CNN) for sentiment analysis

*Convolution Neural Network* (CNN) is one of the most popular deep learning models. Given in figure 1 is the general architecture of such CNN system. The first layer builds the vector from the words in the sentence. Input documents are transformed into a *matrix*, each row of which corresponds a word in a sentence. For example, if we have a sentence with 10 words, each word was represented as a *word-embedding* [19] vector of 100 dimensions, the matrix will have the size of 10x100. This is similar to an image with 10x100 pixels. The next layer will perform *convolution* on these vectors with different filter sets and then *max-pooling* is performed for the set of filtered features to retain the most important features. Then, these features are passed to a fully connected layer with softmax function to produce the final probability output. *Dropout* [20] technique is used to prevent overfitting.

In [16], basic steps of using a CNN in sentiment analysis was detailed in the process by which one feature is extracted from one filter as follows.

Given a sentence with *n* words, let $x_i \in R^k$ be e *k*-dimensional word vector corresponding to the *i*-th word in the sentence. The sentence can be represented as:

$$x_{i:n} = x_1 + x_2 + \dots x_n$$

Here, + denotes vector concatenation. Generally, $x_{i:i+j}$ represents the vector from index $i$ to $i + j$. A convolution operator with filter w $\in R^{h \times k}$ for h words will produce the feature:

$$c_i = f(W . x_{i:i+h-1} + b),$$

Here, $b$ is the bias and $f$ is a non-linear function. By applying the filter on all windows of the sentence, we will obtain the feature map:

$$c = [c_1, c_2, \ldots, c_{n-h+1}].$$

The max-pooling is applied over the feature map and get the maximum value $\hat{c} = \max\{c\}$ as the feature corresponding to this filter.

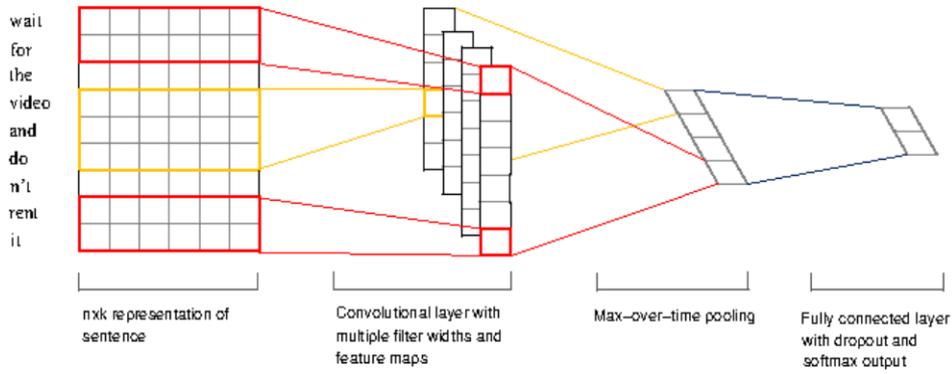

**Fig. 1.** Using CNN for text processing [16]

## 2.2 Using domain knowledge for sentiment analysis

In general, when one performs sentiment analysis for a particular domain, the domain knowledge can be applied as shown in figure 2.

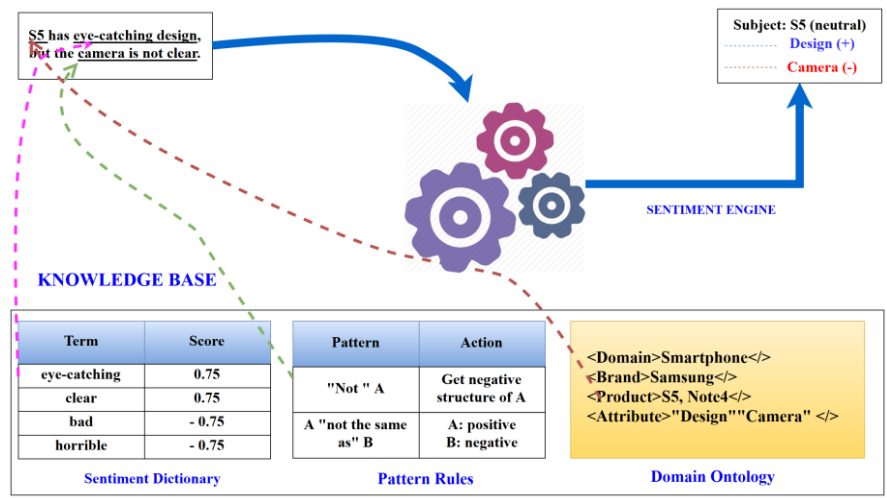

**Fig. 2.** Applying domain knowledge for sentiment analysis.

Thus, a general system [21] will rely on a *Sentiment Engine* to perform sentiment analysis on a user's comment expressing his opinion. This *Sentiment Engine* will operate based on a *Knowledge Base* consisting of the following components:
- A sentiment dictionary, including the *positive* and *negative* sentiment terms. In particular, those sentiment words will be assigned numerical scores indicating their *sentiment levels*.
- Linguistic patterns used to identify different phrase samples.
- A *Sentiment Ontology* to manage semantic relationships between sentiment terms and domain concepts. For more detail of Sentiment Ontology, please refer to [21].

Obviously, determining the sentiment scores for those sentiment terms is an important task to let such a system operate efficiently. We will present this work in the later part of the paper.

## 3 The Proposed Deep Architecture

Figure 3 presents an overview of our proposed deep architecture for sentiment analysis. The system includes the following modules:

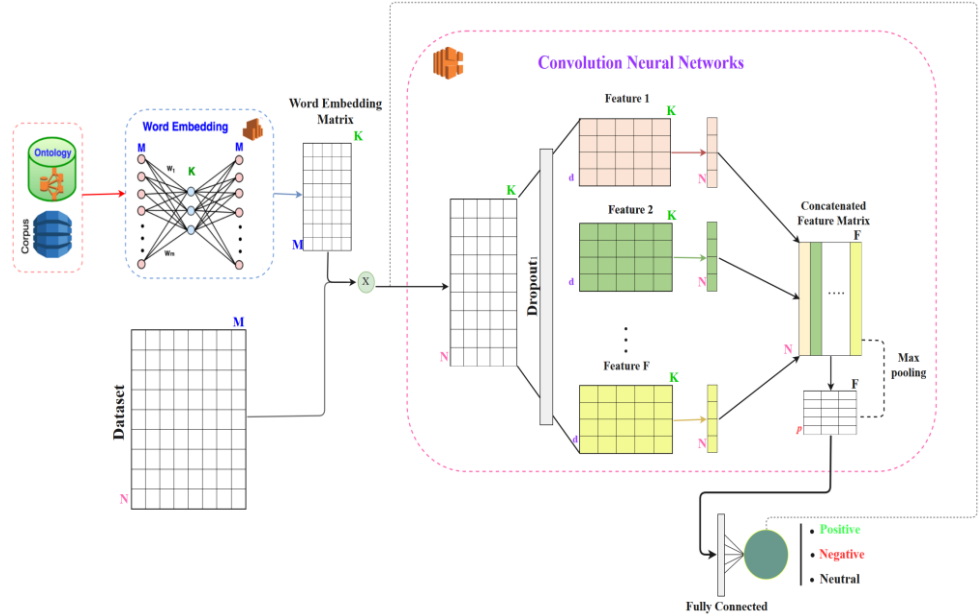

**Fig. 3.** The overall deep architecture.

*Word embedding module*: This is a three-layer neural network $W$. The input layer consists of $M$ words where $M$ is the number of words in the dictionary. The hidden layer consists of *K* nodes with *K* being quite small compared to *N*. The output layer also includes *M* nodes. This network will be trained from an *M*-word dictionary. Each word *w* in the dictionary is passed to the input layer of *W* as a *one-hot vector* corresponding to *w*. The *W* network will be trained to recognize the words *w'* close to *w*, to be activated by the corresponding nodes in the output layer. *w'* words can be determined from a predefined domain ontology built by the expert or learned from a co-occurrence between (*w*,*w'*) in a large corpus in the domain being handled. After *W* is trained, the $w_{ij}$ weights from the input node *i* to the intermediate node *j* will form a word embedding matrix $W_{M \times K}$.

*Training dataset*: A set of collected documents. Each of these documents has been labeled {*positive, neutral, negative*} w.r.t to an object that needs to be sentiment-rated. A document with *N* words will be represented as matrix $D_{N \times M}$, in which the $i^{th}$ row is the one-hot vector corresponding to the $i^{th}$ word in the document. When performing the matrix multiplication $D \times W$, we can obtain an embedded matrix $E_{N \times K}$. Matrix *E* will be used as the input for next *Convolution Neural Network* module.

*Convolution Neural Network:* At this stage, the matrix *E* will be collapsed with a convolutional window, which is a matrix $F_{d \times K}$. The meaning of this matrix *F* is to extract an *abstract feature* from the *d-gram* analysis of the original text. The system will use *f* matrices $F_{d \times K}$ as an attempt to learn *f* abstract features. With the convolution between two matrices *E* and *F* being a *N*×1 column matrix, we will obtain the last matrix $C_{N \times f}$ by concatenating these column matrices together.

Next, matrix *C* will be fed into a pooling layer by a window *p*×*f*. The meaning of this process is to keep the important *d-gram* sets in consecutive *p d-gram*. Finally, we obtain the matrix $Q_{q \times f}$ with *q = N/p*.

Finally, a fully connected layer will be implemented to aggregate the results and from there conducted back-propagation process.

## 4 Sentiment Scores Learning

### 4.1 Problem definition

In this section, we discuss on learning sentiment score. In a general sentiment system, an adverb is used to modify or qualify a sentiment word. Mathematically, each sentiment word is associated with a sentiment score. If this sentiment word is associated with an adverb, the sentiment score would be scaled by the adverb's score. For example, consider the text "*S5 is very beautiful*". The final score of the entity S5 (1.125) is calculated by multiplying the score of the sentiment word "*beautiful*" (0.75) with the adverb "*very*" (1.5). Since it is a positive number, one can conclude that this text is a positive mention to the product *S5*.

In general, the final score of a mention is a linear combination of all pair of sentiment word and its associated adverb as follows:

$$f = \sum_{i=1}^{N} a_i s_i = a^T s \qquad (1)$$

where ($a_i$, $s_i$) is a pair of (adverb-sentiment word) in the text; $a_i$ is set as 1 for every sentiment word $s_i$ that does not have any associated adverb.

**4.2 Sentiment score learning**

**4.2.1 Adverb score learning**

For each mention in a training set of $M$ mentions, let $t^{(m)}$ be the score of the $m^{th}$ mention, $s^{(m)}$ the vector form of the sentiment words, and $b^{(m)}$ the bias (calculated by sum of all sentiment words $s_i^{(m)}$ that do not have any associated adverb). We would like to find the set of adverb's scores $w$ that minimizes the error:

$$\arg_w \min E(w) = \sum_{m=1}^{M} [(w^T s^{(m)} + b^{(m)} - t^{(m)})]^2 \qquad (2)$$

The above equation leads to a traditional least square problem. In the context of sentiment analysis, in addition to minimizing the error function $E$, we also do not want $w_i$ to take a negative value. Furthermore, the norm of $w$ should not to be large (which leads to an overfitting). So, the quadratic optimization problem for adverb score learning is formulated as:

$$\arg_w \min \sum_{m=1}^{M} [(w^T s^{(m)} + b^{(m)} - t^{(m)})]^2 + \lambda w^T w \qquad (3)$$

$$\text{subject to } w >= 0$$

where $\lambda$ is the regularization parameter which stands for the trade-off between error minimization purpose and the overfitting avoidance.

**4.2.2 Sentiment word score learning**

This process is similar to the adverb score learning problem, but we consider the set of adverb scores $w$ as fixed. We denote $\theta^{(m)}$ as scale parameters which are associated with the sentiment word set $s$; and set $\theta_i^{(m)} = 1$ if the sentiment word $s_i$ does not pair with any adverb in the $m^{th}$ mention, otherwise $\theta_i^{(m)} = w_j^{(m)}$. Moreover, to model two types of sentiment word (positive and negative), we constrain $s_i > 0$ or $s_i < 0$ depending on whether the $i^{th}$ sentiment word is positive or negative. The optimal set of sentiment words $s$ is the solution of the following quadratic optimization problem:

$$\arg_s \min \sum_{m=1}^{M} (\theta^{(m)^T} s - t^{(m)})^2 + \lambda s^T s \qquad (4)$$

$$\text{subject to } As < 0$$

where $A$ is the diagonal matrix such that $A_{i,i} = \pm 1$, indicating whether $s_i$ is positive or negative sentiment word.

### 4.2.3 Iterative learning

In this section, we combine the learning process in 4.2.1 and 4.2.2 together to form a dictionary learning algorithm that iteratively trains both the adverbs $a$ and the sentiment word $s$.

---
**Algorithm 1** Iterative Sentiment Word Learning
---
1: **procedure** TRAIN
2:    $K \leftarrow$ max iterations
3:    $s \leftarrow$ load sentiment words from database
4:    Initialize $b^{(m)}, t^{(m)}, s^{(m)}$ from *training set*
5:    Initialize $\lambda, A$
6:    $k \leftarrow 0$
7: loop for $k < K$:
8:    $w \leftarrow$ solving adverb optimization problem in (3)
9:    Update $\theta^{(m)}$ from $w$
10:   $s \leftarrow$ solving sentiment word optimization problem in (4)
11:   Update $s^{(m)}$ from $s$
12:   $k \leftarrow k + 1$
---

### 4.2.4 Using learned sentiment scores for data augmentation

*Data augmentation* is a technique commonly used in learning systems to increase the size of training data sets as well as to control generalization error for the learning model by creating different variations from the original data. For example, for image processing, one can re-size an image to generate different variants from this image. In our case, from a dataset of labeled samples, we will generate variants by replacing the sentiment terms in the original data by the other sentiment terms that *have similar absolute score*s.

For example, let us consider an emotional sentence "*Company A is better than Company B. Company B is horrible*", with the object that needs to be analyzed being *Company B*, the system will first preprocess the sentence as *"Company A is better than **Target**. **Target** is horrible."* Obviously, this sentence will be labeled as negative, w.r.t **Target**.

In this example, we assume that the words such as *horrible*, *poor*, *terrible* have similar negative scores after our learning process. In addition, the words *great* and *amazing* have similar absolute values of opposite sign (i.e these words have positive scores). Thus, from this sample, we will generate other augmented training samples as follows.

| # | **Training Data** | **Label** |
|---|---|---|
| 1 | *Company A is better than **Target**. **Target** is poor.* | Negative |
| 2 | *Company A is better than **Target**. **Target** is terrible* | Negative |
| 3 | *Company A is worse than **Target**. **Target** is great.* | Positive |
| 4 | *Company A is worse than **Target**. **Target** is amazing* | Positive |

In our learning system, the generation of augmented positive samples from the original negative samples is important, as this will help the system recognize that the word *Company A* does not play any role in identifying emotions since it appears in both positive and negative samples. Conversely, sentiment orientation will be determined by the sentiment words, including the original words and newly replaced words.

## 5  Using Penalty Matrix for the Loss Function

In neural network systems, one of the common methods for evaluating the loss functions is *cross entropy* [22]. Generally, a mention sample will be labeled with a 3-dimensional vector *y*. Each dimension respectively represents a value in [*positive, negative, neutral*]. For example, if a mention is labeled as negative, the corresponding *y* vector of this mention is (0,1,0). After the learning process, a vector of probability distribution over labels of 3-dimensional $\bar{y}$ will be generated, corresponding to the learning outcome of the system. The loss function is then calculated by the cross entropy formula as follows:

$$H(y,\bar{y}) = -\sum y_i \ln(\bar{y}) \qquad (5)$$

However, unlike standard classification task, the importance of each label in sentiment analysis (*positive*, *negative*, *neutral*) is different. Generally, in this domain, the data is unbalanced. That is, the number of *neutral* mentions are very large, as compared to other labels. Therefore, if a mention is classified as *neutral*, the probability that it is a misclassified case is lower than the case it is classified as *positive/negative*. Moreover, positive and negative are two distinctly opposite cases. Thus, the error punished when a mention, expected as *neutral*, is misclassified as *positive*, should be less than that of the case where a *negative*-expected mention is misclassified as *positive*. The loss function is calculated by the default cross entropy function does not reflect those issues. Thus, we introduce a custom loss function, known as *weighted cross entropy* in which the cross entropy loss is multiplied by a corresponding penalty weight specified in a *penalty matrix*:

Table 1. The penalty matrix

| predicted/expected | positive | negative | neutral |
|---|---|---|---|
| positive | 1 | 4 | 3 |
| negative | 4 | 1 | 3 |
| neutral | 2 | 2 | 1 |

According to the penalty matrix in Table 1, one can observe that if a mention is expected to be *negative* but is predicted as *positive* or vice versa, the corresponding penalty weight is 4. Meanwhile, for the case that a mention is expected to be *positive* or *negative* and predicted as *neutral*, the penalty weight is 2. In other words, the former case is considered more serious than the latter. Also, if a mention is expected to be *neutral* and predicted as *positive* or *negative*, the penalty weight is 3. It is obvious that if the prediction and the expectation match to each other, the loss is not weighted as the penalty weight value is 1 (i.e the loss function will be minimized in this case).

**Example 1.** If y is [0,1,0] (*negative*), $\bar{y}$ = [0.2,0.3,0.5] (*neutral*), then the default cross entropy will result in 1.204, while the result of weighted cross entropy is 2.408.

**Example 2**. If *y* is [1,0,0] (*positive*), $\bar{y}$ = [0.2,0.7,0.1] (*negative*), the default cross entropy will also result in 1.609, while the result of weighted cross entropy is 6.436.

Example 1 and Example 2 show that the weighted cross entropy function gives different loss values to different misclassification cases. Currently, we develop our penalty matrix based on observable intuition. However, in the future, we can rely on the distribution of data to construct this penalty matrix.

## 6  Experimental Results

We have applied our enhancement on basic deep learning model for sentiment analysis. The data we collected included 1 million social network discussions with labeled sentiment. This dataset is provided by YouNet Media[1], a company that analyzes data on social media channels. The company also provides a set of initial sentiment dictionary that includes positive and negative terms. However, these sentiment terms are manually assigned by sentiment scores of only 4 values in (1.0.5, -0.5, -1).

Initially, the data were represented as one-hot vectors consisting of 65000 dimensions. After performing the word embedding technique, these vectors were reduced to 320 dimensions. In the Convolution Neural Network, we then used 128 filters. For training, we applied *k*-fold cross validation strategy with $k = 5$. Since our data is unbalanced between positive, neutral and neutral samples, we use the SMOTE [23] sampling method to balance data.

Besides CNN network model, we also employ the traditional SVM classification method using a *bag-of-word* approach for testing. In our experiment, we enhance the original CNN model with our improvements. In the CNN-*quad* method, we use quadratic programming to learn sentiment scores, instead of using default values in the sentiment dictionary. In the CNN-*cross* method, we use weighted cross entropy to calculate the loss function. Finally, the CNN-*total* method combines two enhancement of data augmentation and weighted cross entropy.

---

[1] http://www.younetmedia.com/

Table 2. Experimental results.

|  | Recall | Precision | *F*-measure |
|---|---|---|---|
| SVM | 81.49% | 75.49% | 78.38% |
| CNN | 88.32% | 85.46% | 86.87% |
| CNN-quad | 91.07% | 90.18% | 90.62% |
| CNN-cross | 87.17% | 93.55% | 90.25% |
| CNN-total | 90.33% | 95.26% | 92.73% |

We use the metrics in information retrieval, including recall, precision, and F-measure to evaluate performance. The results showed that CNN-based methods achieved better performance than the traditional SVM method. One can also observe that using quadratic programming to calculate sentiment scores for data augmentation has significantly increased recall and precision.

Finally, the use of weighted cross entropy slightly reduces recall, but it makes precision increased significantly, as the system learns better from serious misclassification such as from negative to positive (and vice versa). Finally, the combined CNN-*total* method yields the best results, in terms of F-measure. This demonstrates the advantage of our approach.

# 7 Conclusion

This paper proposed an approach to improve the accuracy of deep learning for sentiment analysis by incorporating domain knowledge. We introduce two improvements, including using quadratic programming to learn the sentiment score for data augmentation and using weighted cross entropy with penalty matrix as an enhanced loss function. When experimented with real datasets, our proposed approach demonstrated significant improvement on the F-measure metric.

**Acknowledgments.** We are grateful to YouNet Media for supporting real datasets for our experiment.